\documentclass[10pt, a4paper]{article}

\usepackage[final]{lrec2026} 

\usepackage{booktabs}
\usepackage{multirow}
\usepackage{longtable}
\usepackage{threeparttable}

\usepackage{enumitem}
\usepackage{makecell}

\title{MiniLingua: A Small Open-Source LLM for European Languages}

\name{Anna Aksenova\thanks{* Equal contribution.}\textsuperscript{*}, Boris Zverkov\textsuperscript{*}, Nicola Dainese, \\ Alexander Nikitin, Pekka Marttinen} 

\address{Aalto University \\
         pekka.marttinen@aalto.fi}

\abstract{
Large language models are powerful but often limited by high computational cost, privacy concerns, and English-centric training. Recent progress demonstrates that small, efficient models with around one billion parameters can deliver strong results and enable on-device use. This paper introduces MiniLingua, a multilingual open-source LLM of one billion parameters trained from scratch for 13 European languages, designed to balance coverage and instruction-following capabilities. Based on evaluation results, the instruction-tuned version of MiniLingua outperforms EuroLLM, a model with a similar training approach but a larger training budget, on summarization, classification and both open- and closed-book question answering. Moreover, it remains competitive with more advanced state-of-the-art models on open-ended generation tasks. We release model weights, tokenizer and source code used for data processing and model training. 
 \\ \newline \Keywords{LLM, pre-training, supervised finetuning} }

\begin{document}

\maketitleabstract

\section{Introduction}
Large language models (LLMs) have demonstrated remarkable natural language processing and general-purpose reasoning capabilities, becoming widely used in the industry. However, several limitations hinder further research progress. First, many leading LLMs are proprietary, which makes it difficult for researchers and practitioners to reproduce results or adapt them to their own needs. Second, training data and benchmarks are still dominated by English, leading to weaker support for many other languages. Third, large models with tens or hundreds of billions of parameters require high computational cost, which limits their usability in resource-constrained settings.

\begin{figure*}[h]
    \centering
    \includegraphics[width=\textwidth]{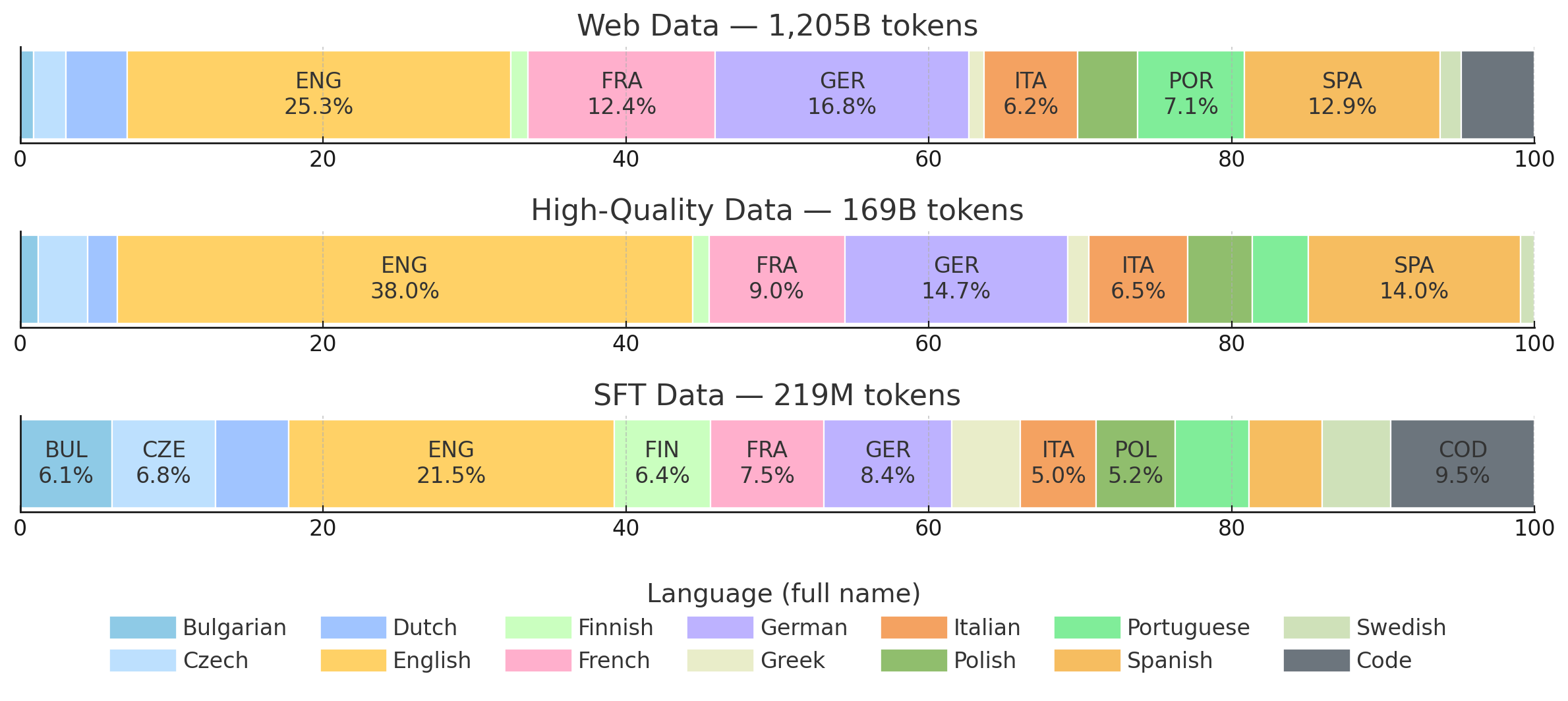}
    \caption{Data distribution per dataset after preprocessing. For all datasets English remains the dominant language, but never exceeds 40\% of the data.}
    \label{fig:data_distribution}
\end{figure*}

To address these gaps, we introduce \mbox{\textbf{MiniLingua}}, a multilingual small language model trained from scratch with approximately one billion parameters. MiniLingua is designed to support 13 European languages (full list is presented in Figure \ref{fig:data_distribution}), enabling better access for languages that are often underrepresented. The model is compact enough to allow on-device use, while still being trained with methods that enable instruction-following abilities. 

\paragraph{Contributions.} The main contributions of this work to the development of multilingual open models development are:
\begin{itemize}[noitemsep, topsep=0pt]
    \item Base and instruction tuned versions of open-source multilingual LLM for 13 European languages, trained from scratch with one billion parameters.\footnote{\url{https://huggingface.co/minilingua-ai}}
    \item Fully open source code for data cleaning, pre-training, fine-tuning, and evaluation.\footnote{\url{https://huggingface.co/minilingua-ai/MiniLingua-1b-Instruct}}
    \item Curated multilingual datasets with permissive licenses for research use.\footnote{\url{https://huggingface.co/datasets/minilingua-ai/mcqa-minilingua-sft}} \footnote{\url{https://github.com/MiniLingua-ai/training_artifacts/tree/main/data}}
\end{itemize}

The remainder of this paper is structured as follows. Section~\ref{sec:related} reviews related work, followed by the description of our datasets, preprocessing pipeline, and model training procedure (Sections~\ref{sec:data}–\ref{sec:sft}). Section~\ref{sec:evaluation} presents the evaluation setup and results. We conclude with a discussion of the main results and limitations (Sections~\ref{sec:conclusion}–\ref{sec:limit}).

\section{Related Work}
\label{sec:related}

LLMs have predominantly been trained with a focus on English; however, recent efforts have led to the emergence of more multilingual models. Some of these models are adaptations of existing monolingual architectures. This approach works best when languages use similar scripts, while distant scripts or rich morphology often reduce transfer quality \citep{conneau2020unsupervisedcrosslingualrepresentationlearning, malkin-etal-2022-balanced}. Other works instead focus on training models on multilingual corpora~\citep{martins2024eurollmmultilinguallanguagemodels}.

In continued pre-training, tokenization remains a persistent challenge. Subword methods such as BPE and SentencePiece~\citep{sennrich-etal-2016-neural, kudo2018sentencepiecesimplelanguageindependent} often split words in morphologically rich or non-Latin-script languages into several pieces, leading to longer sequences, higher costs, and uneven performance across languages~\citep{rust-etal-2021-good, velayuthan2024egalitarianlanguagerepresentationlanguage}. 

Training the model from scratch, on the other hand, allows for training a tokenizer on a multilingual corpus, facilitating more balanced language representation from the early stages.

For European languages, several small multilingual models have been introduced. LLaMA 3.2-1B~\citep{grattafiori2024llama3herdmodels} and Granite-1B~\citep{granite2024granite} primarily serve high-resource languages. Aya-8B~\citep{dang2024ayaexpansecombiningresearch}, EMMA-500 \mbox{\citep{ji2025emma500enhancingmassivelymultilingual}}, and Apertus~\citep{hernándezcano2025apertusdemocratizingopencompliant} extend coverage, though with uneven quality or larger parameter count. The recent EuroLLM project~\citep{martins2024eurollmmultilinguallanguagemodels} offers the most comprehensive EU coverage, with a tokenizer specifically adapted to European linguistic diversity. However, it is primarily designed for machine translation. In contrast, our work introduces a small multilingual model designed for instruction following, a direction largely unexplored for small model scale and European languages.  

Finally, some high-performing multilingual models have recently been released by industry. Gemma-3-1B \cite{gemmateam2025gemma3technicalreport}, developed by Google DeepMind, is trained on an undisclosed 2T-token dataset. It heavily relies on knowledge distillation from larger models during both pre-training and fine-tuning, and is subsequently aligned using reinforcement learning from human feedback (RLHF)~\cite{ouyang2022training} and additional automated feedback for mathematical and reasoning tasks.
SmolLM2-1.7B \cite{allal2025smollm2smolgoesbig}, from Hugging Face, is instead trained on a corpus of 11T tokens through a multi-stage pre-training process, followed by instruction tuning and RLHF; both the dataset and the code are open-sourced. While these industry-developed models represent the current standard, they are either partially closed, or require substantial computational resources. Our work focuses instead on open, reproducible multilingual modeling within academic resource limits.

\section{Data}
\label{sec:data}

\looseness=-1 To train MiniLingua, we used several datasets and data sources. Generally, they could be grouped as web pre-training data, high-quality pre-training data, and data for supervised fine-tuning. The relative data distribution for each of these datasets is presented in Figure~\ref{fig:data_distribution}. Although English still accounts for the largest part of the data, it never exceeds 40\%, which is significantly smaller compared to other models with which we compare (e.g. 50\% for EuroLLM).

\subsection{Pre-training data}

\begin{figure}[h]
    \centering
    \includegraphics[width=0.8\linewidth]{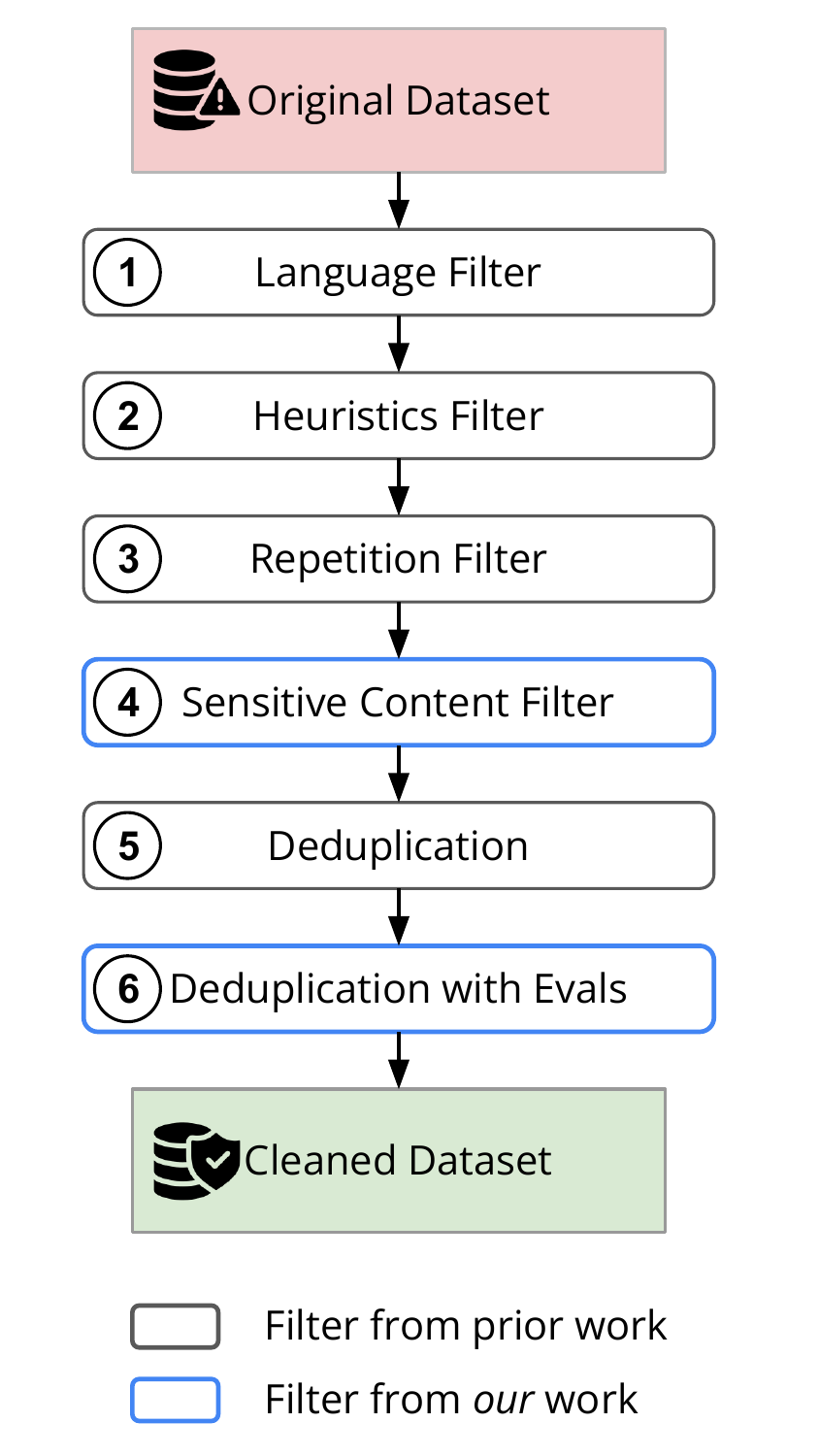}
    \caption{Data cleaning pipeline steps. Steps 4 and 6 were applied to Web dataset, all steps were applied to high-quality data, steps 1 and 6 were applied to SFT data.}
    \label{fig:data_pipeline}
\end{figure}

For general \textbf{web data}, we selected the FineWeb-2 corpus \citep{penedo2024}, a comprehensive and high-quality dataset covering more than 1000 languages. This corpus outperformed other multilingual datasets such as CC-100 \citep{wenzek-etal-2020-ccnet}, mC4 \citep{xue-etal-2021-mt5}, CulturaX \cite{nguyen-etal-2024-culturax}, and HLPT \citep{degibert2024newmassivemultilingualdataset} in various evaluation tasks \citep{fineweb-2}. The FineWeb-2 dataset was derived from 96 Common Crawl snapshots, collected from summer 2013 through April 2024, processed using the Datatrove data processing library, resulting in approximately 8 terabytes of compressed text containing nearly 3 trillion words \citep{fineweb-2}.

In addition, we manually collected \textbf{high-quality datasets} – news, books, subtitles, encyclopedias, government documents and speech transcripts -- for each language in proportions that match our language distribution for training. The source attribution of those datasets is presented in Appendix~\ref{appendix}.

The \textbf{code data} was instead sourced from The Stack v2, a dataset with more than three billion files covering more than 600 programming languages, derived from the Software Heritage archive \cite{lozhkov2024starcoder2stackv2}. The dataset facilitates training in code generation and programming reasoning.



All the datasets that were used for model training had to go through various cleaning steps depending on the data source. The full flow of the \textbf{data cleaning pipeline} is shown in Figure~\ref{fig:data_pipeline}. It includes:

\begin{enumerate}
    \item Language filter with Fasttext language identification models \citep{joulin2016bag}.
    \item A set of heuristic-based filtering such as too short or too long documents, percentage of bullet lines, percentage of non-alphanumerical characters etc. suggested by \citet{rae2022scalinglanguagemodelsmethods}.
    \item Repetition filter for documents with highly repetitive symbols, words or n-grams \citep{rae2022scalinglanguagemodelsmethods}.
    \item Use of a multilingual blacklist of inappropriate terms, built from \href{https://github.com/LDNOOBW/List-of-Dirty-Naughty-Obscene-and-Otherwise-Bad-Words}{LDNOOBW}. In addition, we translated the list of bad words from datatrove \citep{penedo2024datatrove} using Google Translate and manually extended and reviewed it with the help of native speakers.
    \item Deduplication of documents with high line or paragraph overlap \citep{rae2022scalinglanguagemodelsmethods}.
    \item Deduplication across training and evaluation datasets using Jaccard similarity to prevent data leakage and ensure fair downstream assessment.
\end{enumerate}

For FineWeb 2, only the obscenity filter and evaluation data deduplication steps were applied due to the fact that it was rigorously cleaned beforehand. The high quality dataset went through the entire cleaning pipeline, while the Stack went only through the last step.

The pipeline had to be adjusted for different languages following the resulting filtering percentage of the data. Overall, we aimed at keeping 80\% of the collected dataset as that was the share we obtained for English data following Gopher guidelines \citep{rae2022scalinglanguagemodelsmethods}. More details on preprocessing can be found in the released codebase.

\subsection{SFT Data}

Because MiniLingua is designed for multilingual and multimodal instruction following, we combined several public instruction-tuning datasets with curated multilingual QA resources to ensure broad coverage and diverse answer formats. From large-scale sources, we used Aya \citep{singh2024ayadatasetopenaccesscollection}, Bactrian-X \citep{li2023bactrianxmultilingualreplicableinstructionfollowing}, and LIMA \citep{zhou2023limaalignment}, while for code reasoning we adopted Self-OSS-Instruct Code 50k \citep{wei2024magicoderempoweringcodegeneration}. 

In addition to publicly available instruction-tuning corpora, \emph{we curated a multilingual question-answering (QA) dataset} to better support instruction following in multiple languages and diverse answer formats. We sourced QA examples from a range of datasets, including grade-school science questions, frequently asked questions, and more. Source datasets are listed in Appendix~\ref{appendix:additional_data_details}.

Based on these sources, we used GPT-4o to \emph{generate instruction-response pairs} with various prompting strategies. For each QA item, we generated instruction-tuning examples by varying \emph{four main factors}: the instruction language (20\% English vs 80\% the target language), instruction variation (e.g., rewordings), expected answer format (letter, number, or full answer), and output formatting style. This resulted in a wide set of naturalistic and linguistically diverse instruction formats.

\looseness=-1 Each SFT training instance was wrapped as follows:
\begin{itemize}
    \item The entire sample begins with the \texttt{<|start\_of\_sequence|>} token.
    \item The instruction section is enclosed between \texttt{<|im\_start|>} and \texttt{<|im\_end|>}.
    \item If a separate input (i.e., context or supporting information) is present, it follows the instruction text directly before \texttt{<|im\_end|>}.
    \item The output (reference response) appears immediately after \texttt{<|im\_end|>} and continues until the \texttt{<|end\_of\_sequence|>} token.
\end{itemize}

For example:
    \begin{center}
    \texttt{<|start\_of\_sequence|><|im\_start|>} \textit{Instruction} \texttt{<|im\_end|>} \textit{Response} \texttt{<|end\_of\_sequence|>}
    \end{center}

Instances that exceeded the maximum sequence length of the model were filtered out. This includes both excessively long instructions and completions, which would otherwise be truncated or split inconsistently during training, potentially degrading model performance.

We applied additional heuristics to detect and remove examples that were too short to be meaningful (e.g., one-word replies or instructions like "Say hi"). Language detection was also performed to ensure that both the instruction and response matched the intended language tag. Pairs with mismatches or low classifier confidence (class logit probability) were discarded to avoid introducing noisy supervision \cite{grave2018learning}.


\section{Tokenizer}
\label{sec:tokenizer}


\looseness=-1 To ensure that tokenization is optimized for our target set of languages, we train a tokenizer from scratch.

\paragraph{Vocabulary size.} We experiment with four vocabulary sizes, 32K, 64K, 96k and 128K, and train tokenizers on cleaned subsets of the FineWeb corpus. The lower end (32K) is typical for monolingual models of moderate size (under 10B parameters), while the higher end (128K) is commonly used in multilingual models, to accommodate the increased diversity in scripts and morphology. Additionally, prior work has shown that having a vocabulary size divisible by 8 can positively influence training stability and optimization performance, due to tensor layout compatibility in common hardware accelerators \citep{tao2024scalinglawsvocabularylarger}. Consequently, we follow this recommendation in our work.

\paragraph{Evaluation metrics.} For evaluation, we use 10\% of the data as a validation set and compute the \textbf{Normalised Sequence Length (NSL)}, defined as the token per word ratio (lower is better). We further consider two multi-lingual variants of it: the \textbf{average NSL}, obtained by averaging per-language NSL scores, and the \textbf{weighted NSL}, where each language score is weighted by its dataset size. The former captures \emph{language-agnostic} performance, whereas the latter reflects tokenizer performance under a \emph{real-world language distribution}.

\paragraph{Data mixtures.} In multilingual applications, selecting the training data for a tokenizer involves a trade-off between maximizing pre-training efficiency and ensuring balanced performance across languages, regardless of their relative representation. We therefore train tokenizers under four data distributions: \emph{Balanced}, where English accounts for only 20\% of the data and no language falls below 3\%; \emph{Intermediate}, where English is still downsampled, but low-resource languages make up at least 1\%; \emph{Original}, which preserves the pre-cleaning language distribution of the web data; and \emph{Train}, which reflects the post-cleaning distribution.

\begin{figure}[h!]
    \centering
    \includegraphics[width=0.5\textwidth]{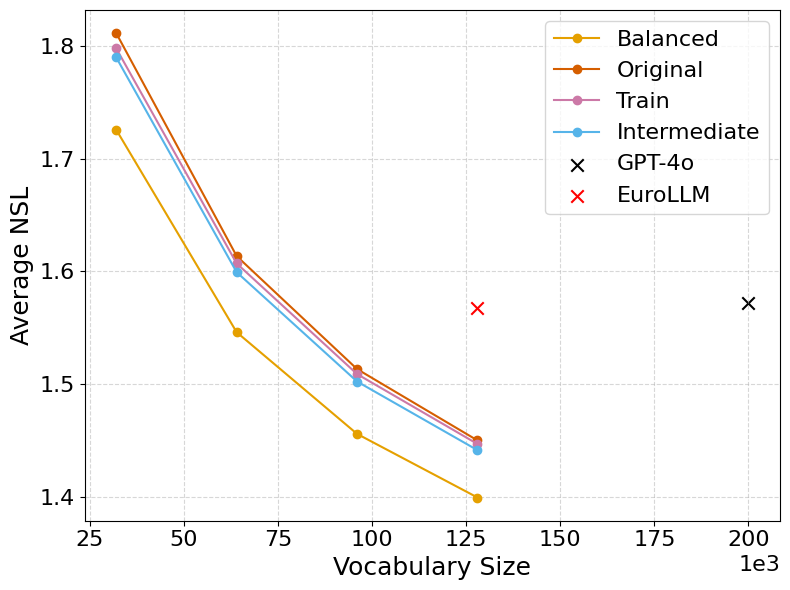}
    \caption{Comparison of NSL for target languages between our tokenizers of varying data mixtures and size with GPT-4o and EuroLLM.}
    \label{fig:avg_nsl}
\end{figure}

\paragraph{Results.} Figure~\ref{fig:avg_nsl} shows the \textbf{average NSL} as a function of vocabulary size for different data mixtures.
Our best tokenizers consistently outperform both GPT-4o and EuroLLM, despite being trained on significantly smaller datasets. Notably, our largest tokenizers achieve lower NSL than EuroLLM across all data mixtures, even though both use a 128K vocabulary. Compared to GPT-4o, which employs a much larger vocabulary \cite{openai_tiktoken}, our 128K models still deliver better compression for all languages considered, except code and English. The \textit{Balanced} configuration, in particular, shows strong performance even with only 64K tokens, indicating that proper data mixture design can have a greater effect than vocabulary size alone. Due to its overall superior performance, we adopt the 128K \textit{Balanced} tokenizer for all further experiments.

\begin{figure*}[h!]
    \centering
    \includegraphics[width=\textwidth]{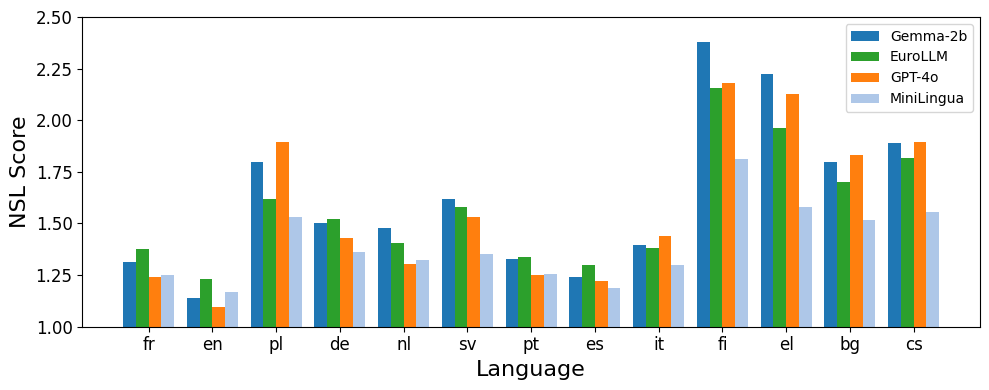}
    \caption{Comparison of NSL per language for MiniLingua (Balanced tokenizer with vocabulary size of 128K), GPT-4o, Gemma and EuroLLM. Our tokenizer shows the most optimal compression rate for the majority of languages without sacrificing performance on English.}
    \label{fig:nsl_per_lang}
\end{figure*}

Figure~\ref{fig:nsl_per_lang} compares the \textbf{NSL across languages} using our trained tokenizer, as well as the tokenizers from EuroLLM, Gemma, and GPT-4o. Overall, our tokenizer achieves the best compression across nearly all evaluated languages. Notably, our tokenizer achieves significantly better compression for lower-resource European languages, particularly Greek, Bulgarian, Finnish, and Czech, while maintaining competitive compression efficiency for English (lower NSL score compared to EuroLLM). These results show that our tokenizer improves coverage for underrepresented languages without compromising performance on high-resource ones.

\section{Pre-training}
\label{sec:pretraining} 

This section outlines the architecture of MiniLingua pre-training. MiniLingua adopts a decoder-only transformer design, following the LLaMA family~\citep{grattafiori2024llama3herdmodels}. The model integrates modern components such as SwiGLU activations, grouped query attention, rotary positional embeddings, and RMSNorm.  

\begin{table}[h!]
\centering
\resizebox{0.5\textwidth}{!}{%
\begin{tabular}{ll}
\toprule
\textbf{Component} & \textbf{Configuration} \\
\midrule
Depth & 32 decoder layers \\
Hidden size & 1536 \\
FFN size & 6144 (SwiGLU) \\
Attention heads & 24 total, 6 key-value (GQA) \\
Positional encoding & Rotary embeddings (RoPE) \\
Vocabulary & 128,008 tokens \\
Embedding & Tied input and output layers \\
Normalization & RMSNorm \\
Precision & \texttt{bfloat16}\\
Optimizer & AdamW, $\beta_1=0.9$, $\beta_2=0.95$, wd=0.01 \\
LR schedule & Warmup–Stable–Decay \\
\bottomrule
\end{tabular}%
}
\caption{MiniLingua architecture and training configuration.}
\label{tab:architecture}
\end{table}

The architecture prioritizes depth over width, aligning with recent findings that deeper small models generalize better~\citep{liu2024mobilellmoptimizingsubbillionparameter, allal2025smollm2smolgoesbig}. Grouped Query Attention (GQA) reduces inference costs while maintaining performance \citep{ainslie2023gqatraininggeneralizedmultiquery, shazeer2019fasttransformerdecodingwritehead}. Rotary Position Embeddings (RoPE) embed positional information directly into queries and keys, improving extrapolation \citep{su2023roformerenhancedtransformerrotary}. Shared input-output embeddings lower memory use \citep{press2017using}, and RMSNorm stabilizes training at scale \citep{zhang2019rootmeansquarelayer, touvron2023llama2openfoundation}. Training is carried out in \texttt{bfloat16} precision, using AdamW with gradient clipping. It consists of three consecutive phases: a warmup phase, a stable phase, and a decay phase \citep{loshchilov2019decoupledweightdecayregularization, hägele2024scalinglawscomputeoptimaltraining, bergsma2025straightzerolinearlydecaying}.  

\subsection{Small Scale Experiments}

We conducted two series of small-scale experiments for estimating the expected loss and effective learning rate and batch size.

We trained three smaller autoregressive transformer models of approximately 30M, 60M, and 110M non-embedding parameters on a representative multilingual web corpus. These models were architecturally aligned with our target 1B parameter model, differing only in depth and width. The detailed configurations are provided in Table~\ref{tab:scaling_configs} of the appendix.

To approximate the scaling law, we fitted a bi-variate power-law function to the observed validation losses as a function of model size \(N\) (non-embedding parameters) and dataset size \(D\) (tokens seen during training) \citep{hu2024minicpmunveilingpotentialsmall}. The fitted scaling law is defined as:

\begin{equation}
\mathcal{L}(N, D) = L_{\infty} + A_N \cdot N^{-\alpha} + A_D \cdot D^{-\beta}
\end{equation}

where:
\begin{itemize}[noitemsep, topsep=0pt] 
    \item \( \mathcal{L}(N, D) \) is the predicted loss for model size \(N\) and dataset size \(D\),
    \item \( L_{\infty} \) is the irreducible loss for an infinite-size model and infinite dataset,
    \item \( A_N, A_D \) are scaling coefficients for model and data dimensions, respectively,
    \item \( \alpha, \beta \) are the corresponding scaling exponents.
\end{itemize}

\begin{figure}[t!]
    \centering
    \includegraphics[width=\linewidth]{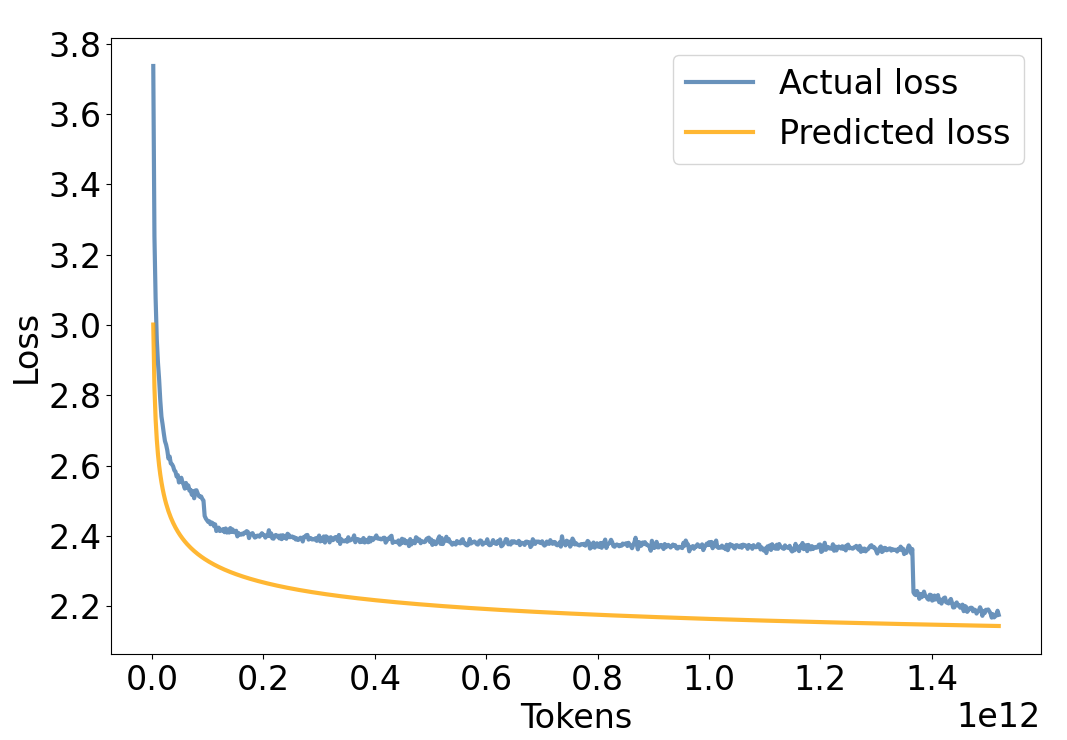}
    \caption{Actual training loss and loss predicted by scaling law formula for one billion size model.}
    \label{fig:scaling_law}
\end{figure}

Fitting the scaling law requires training multiple smaller models and evaluating their performance as a function of training FLOPs (Appendix~\ref{app:scale}). The loss curve for the 1B model, predicted via the scaling law, is presented in Figure~\ref{fig:scaling_law}. The results of the scaling law experiments are consistent with the observed training loss. However, given the relatively small model size and limited compute budget, the accuracy of the fit could be improved by training a larger number of models with greater compute.

The second small-scale experiment series was conducted to determine optimal learning rate and batch size configurations for models of varying parameter sizes. This follows recent recommendations to explore hyperparameters on small models \citep{zhang2022optopenpretrainedtransformer} and to jointly optimize model and training setup under a fixed compute budget \citep{hoffmann2022trainingcomputeoptimallargelanguage}.

Each model was trained using a constant per-device local batch size that optimizes GPU utilization. The global batch size (GBS) was then increased by distributing training over multiple GPUs. We explored global batch sizes ranging from 32K to 2M tokens, corresponding to 1 to 32 nodes in a distributed training setting. Learning rates were varied across a grid: $0.0005, 0.001, 0.0025, 0.004, 0.005$. Validation loss was measured after a fixed training budget for each configuration. 

\looseness=-1 As a result, global batch size of 2M tokens and learning rate of $5 \times 10^{-5}$ for the stable stage were selected.

\subsection{Training Run}

\looseness=-1 The pre-training phase of our language model lasted approximately 12 days and was executed on 32 nodes with four AMD MI200X GPUs each from the LUMI supercomputer \footnote{\href{https://www.lumi-supercomputer.eu/about-lumi/}{https://www.lumi-supercomputer.eu/about-lumi/}}
Training was performed using Megatron-LM framework optimized for AMD architecture. Over this period, the model processed 715,000 training iterations, summing up to approximately $1.2 \times 10^{12}$ FLOPs.

We followed a Warmup-Stable-Decay (WSD) \cite{wen2024wsd} learning rate schedule: during the warmup stage, the learning rate was linearly increased to 0.0025 over 6000 training steps. In the stable stage, it was held constant at 0.0025 for 40,000 training steps, after which it was reduced to 0.0005 due to declining improvements in performance. In the final 10\% of pre-training, the learning rate was linearly decayed to zero. 

For the decay stage, we ran some preliminary experiments with different data mixtures. As a result, we adopted a mixture of 30\% of high-quality data, and the remaining share of web data distributed as in the stable pre-training stage. More details on experimental mixtures can be found in Appendix~\ref{app:mix}.

Figure~\ref{fig:loss_training} shows the training and validation loss curves, with the learning rate indicated by the green dashed line. Several gradient spikes appeared during pre-training, but since they did not lead to divergence and training recovered quickly, we considered them benign. The sharp drop in loss observed during the decay phase is due to the upsampling of high-quality data in this stage, naturally resulting in a lower language modeling loss.

\begin{figure*}[h!]
    \centering
    \includegraphics[width=\textwidth]{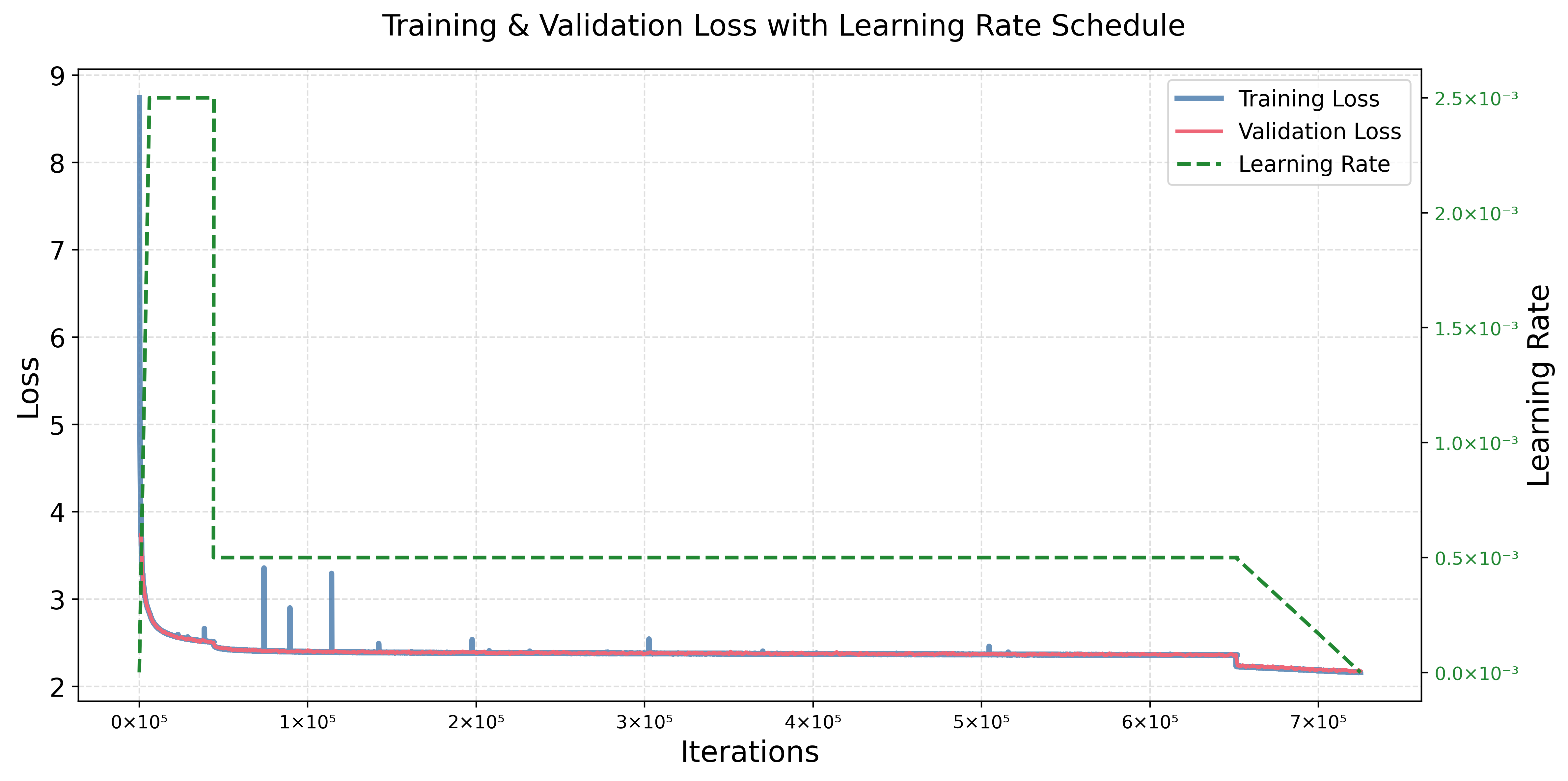}
    \caption{Training and validation loss curves. Loss values were smoothed using a sliding window of 100 steps. Learning rate value corresponding to each iteration is presented in green.}
    \label{fig:loss_training}
\end{figure*}

\section{Supervised Fine-tuning}
\label{sec:sft}

To examine the impact of multilingual imbalance, we experimented with three different sampling strategies during fine-tuning: \textbf{Original distribution} – using the pre-cleaning web data distribution (see Table~\ref{tab:messages_by_language} in Appendix); \textbf{Train distribution} – matching the initial Web Data pre-training proportions (see Figure~\ref{fig:data_distribution}); \textbf{Equal distribution} – uniformly sampling with repetition across all languages for equal token count.

\looseness=-1 After comparison, we selected the \textbf{equal distribution} strategy for the final supervised fine-tuning stage. The final run lasted approximately 25 hours on four H200 GPUs and used one epoch over the equally sampled dataset, using repeated samples from underrepresented languages to balance the corpus. The maximum effective batch size was 256. We fixed the learning rate to $2 \times 10^{-5}$ after preliminary experiments showed that smaller values caused early stagnation, while larger ones led to overfitting. Following common SFT practice, we applied output-token-only loss, masking instruction tokens and training the model solely to predict response tokens.

\section{Evaluation}
\label{sec:evaluation}

We empirically evaluated MiniLingua against two categories of multilingual language models of approximately the same size.

First, we included \textbf{EuroLLM-1.7B}, which follows a similar training setup to ours — pre-training followed by supervised fine-tuning — and was developed under academic settings with a larger overall data and compute budget. This first comparison highlights differences mainly due to dataset composition and basic training methodology.

Second, we compared MiniLingua with industry-developed models \textbf{Gemma-3-1B} \cite{gemmateam2025gemma3technicalreport} and \textbf{SmolLM2-1.7B} \cite{allal2025smollm2smolgoesbig}, from Google DeepMind and Hugging Face respectively, both trained on substantially larger corpora and compute budgets, and aligned with RLHF. This second comparison allowed us to explore the influence of a larger pre-training dataset and the inclusion of an additional alignment stage.

We evaluate both the base and instruction-tuned versions of the models on four benchmarks, and consider an additional one for the instruction-tuned models: \textbf{FLORES-200} \citep{nllbteam2022languageleftbehindscaling}, a 200-language translation benchmark with 3,001 parallel sentences; \textbf{Belebele} \citep{bandarkar-etal-2024-belebele}, a 122-language open-book QA dataset with multiple-choice questions; \textbf{MMLU-X} \citep{thellmann2024crosslingual}, a machine-translated extension of the MMLU benchmark \citep{hendryckstest2021}, covering 57 knowledge domains; \textbf{SIB-200} \citep{adelani2024sib200simpleinclusivebig}, a 200+ language topic classification benchmark from FLORES-200; \textbf{MassiveSumm (MSum)} (for instruction-tuned models only), a multilingual summarization benchmark \citep{varab-schluter-2021-massivesumm} with 28.8M news articles in 92 languages.

\begin{table*}[ht]
\centering
\caption{Base model evaluation across multilingual benchmarks using zero-shot prompting.}
\label{tab:base_model_comparison}
\resizebox{\textwidth}{!}{%
\begin{threeparttable}
\begin{tabular}{l c c c c | c c}
\toprule
\textbf{Model} & \textbf{FLORES $\uparrow$} & 
\textbf{Belebele $\uparrow$} & \textbf{SIB $\uparrow$} & \textbf{MMLU $\uparrow$} & \textbf{Data (tokens)} & \textbf{GPU Hours} \\
\midrule
MiniLingua-1b (ours)    & 0.343 & 0.230 & \textbf{0.248}  & 0.240 & 1.5T & 75K  \\
EuroLLM-1.7b     & 0.360 & 0.250 & 0.230           & 0.250 & 4T   & 500K  \\
\midrule
Gemma-3-1b-pt    & 0.319 & 0.220 & 0.230           & 0.200 & 2T   & $\sim$300K \\
SmolLM2-1.7b     & \textbf{0.367} & \textbf{0.320} & 0.240 & \textbf{0.320} & 11T  & – \tnote{1}\\
\bottomrule
\end{tabular}%
\begin{tablenotes}
\small
\item [1] Training compute is not reported in SmolLM paper.
\end{tablenotes}
\end{threeparttable}
}
\end{table*}

\begin{table*}[ht]
\centering
\caption{Instruction-tuned model evaluation across multilingual benchmarks with English-language prompts.}
\label{tab:sft_model_eval}
\resizebox{0.875\textwidth}{!}{%
\begin{threeparttable}
\begin{tabular}{lcccccc}
\toprule
\textbf{Model} & \textbf{FLORES $\uparrow$} & \textbf{Belebele $\uparrow$} & \textbf{SIB $\uparrow$} & \textbf{MMLU $\uparrow$} & \textbf{MSum $\uparrow$} \\
\midrule
MiniLingua-1b-Instruct (ours) 
& \textbf{0.681} & 0.262 & 0.149 & 0.245 & \textbf{0.187} \\
EuroLLM-1.7b-Instruct 
& – \tnote{2} & 0.216 & 0.124 & 0.205 & 0.0138 \\
\midrule
Gemma-3-1b-it        
& 0.494 & 0.366 & 0.558 & \textbf{0.311} & 0.001 \\
SmolLM2-1.7b-Instruct 
& 0.496 & \textbf{0.369} & \textbf{0.613} & 0.271 & 0.015 \\
\bottomrule
\end{tabular}%
\begin{tablenotes}
\small
\item [2] EuroLLM results on FLORES are omitted since it was trained on the dataset dev set.
\end{tablenotes}
\end{threeparttable}
}
\end{table*}

\paragraph{Evaluation protocol.} For multiple-choice question answering and classification datasets, we evaluated the base model by computing the log-probabilities of the candidate answer tokens. For the instruction-tuned models, we instead compared the generated outputs to the expected answers using fuzzy string matching, allowing variation in whitespace and punctuation, but requiring the exact answer to be present. For open-ended generation tasks, we report COMET scores for translation on FLORES \citep{rei2020cometneuralframeworkmt} and ROUGE scores for summarization on MassiveSum \citep{lin-2004-rouge}.

\paragraph{Results.} The results for \textbf{base model evaluation} are summarized in Table~\ref{tab:base_model_comparison}.
Despite using smaller compute budget and training dataset (up to 5 and 7 times smaller respectively), MiniLingua performs competitively with larger and more compute-intensive models. On FLORES-200, our model achieves a COMET score of 0.343, outperforming Gemma-3-1b and approaching EuroLLM-1.7b. In classification and QA tasks such as SIB-200 and MMLU-X, our model surpasses both Gemma and EuroLLM, with 0.248 and 0.240 accuracy, respectively. While SmolLM2-1.7b remains the strongest model overall due to its significantly larger training dataset (11T tokens), our results suggest that with careful data curation, multilingual balancing, and an optimized training strategy, the base model is competitive with the state of the art.

Table~\ref{tab:sft_model_eval} reports results of \textbf{instruction-tuned models} on multilingual benchmarks using English prompts. 
The strongest overall model in this comparison is SmolLM2-1.7b-Instruct, followed by Gemma-3-1b-it. Both models benefit from training on larger corpora, as well as alignment through RLHF. In addition, Gemma-3-1b-it SFT phase included distillation of a 27b parameter model. In contrast, MiniLingua-1b-Instruct and EuroLLM-1.7b-Instruct were fine-tuned without human feedback, which explains their weaker performance on strict question answering and classification tasks. 

A closer look at task types reveals more nuance. For QA and classification tasks, which require single-letter answers (e.g., Belebele, SIB, MMLU), MiniLingua underperforms compared to aligned models, largely due to incorrect format generation. 
However, for generation tasks that require to produce open-ended answers in the target language, such as translation (FLORES, in English-to-target) and summarization (MassiveSum), MiniLingua achieves the highest score. Instead, Gemma and SmolLM both fail to consistently output the right language, defaulting to English and thus achieving a lower score.

Importantly, when comparing to a model of similar scale and resources such as EuroLLM-1.7b-Instruct, MiniLingua consistently performs better across all tasks. This indicates that our focus on balanced multilingual coverage and careful data preparation leads to better outcomes than scaling alone. In other words, MiniLingua is not designed to outperform highly resourced models like SmolLM or Gemma, but rather to provide an open, efficient, and well-documented alternative that delivers strong results for its size and training budget.

\section{Conclusions}
\label{sec:conclusion}

This work presented the design, development and evaluation of MiniLingua, a multilingual large language model, from data collection and preprocessing to model pre-training and instruction tuning. Even with constrained computational resources, we demonstrated that careful design decisions such as training a balanced tokeniser and curating the dataset can yield competitive results.

Despite its smaller size and reduced training budget, MiniLingua matches or outperforms the larger EuroLLM-1.7B across both base and instruction-tuned settings.

To promote transparency and facilitate further research, we openly release the models, a detailed list of used datasets including licenses, training curves, training configuration files, and evaluation scripts under permissive licenses. We hope that these assets will improve future work on resource-efficient multilingual language models, especially in low- and mid-resource European languages.

\section{Limitations}
\label{sec:limit}

This work has several limitations. First, we did not perform systematic ablation studies on data cleaning strategies or tokenizer vocabulary size. Both were fixed early in the pipeline due to computational constraints. The work was done under tight compute budget of 100K GPU hours, which limited the scope of model variants and experiments we could conduct. Furthermore, our pipeline lacks an explicit alignment stage such as RLHF due to the absence of high-quality multilingual preference data. Furthermore, our evaluations relied largely on translated, English-centric benchmarks, which may introduce cultural bias and reduce generalizability. Finally, we cover only 13 European languages, while there are more languages used in Europe. Future work with broader compute resources and community-driven dataset efforts could help address these limitations.

\section{Acknowledgements}

We thank Yu Tian and Simo Tuomisto for their assistance with the cluster setup, and Risto Luukkonen for sharing his experience and the ROCm adaptation of the Megatron-LM library. We are also grateful to the native speakers who contributed to the analysis of multilingual data.

We acknowledge CSC – IT Center for Science for granting access to the LUMI supercomputer, owned by the EuroHPC Joint Undertaking, and hosted by CSC (Finland) and the LUMI consortium through Finland.

The supervised fine-tuning and data processing were performed using the computational resources of the Triton supercomputer provided by the Aalto University School of Science “Science-IT” project.

This work was supported by the Research Council of Finland (Flagship programme: Finnish Center for Artificial Intelligence FCAI, and grants 352986, 358246) and EU (H2020 grant 101016775 and NextGenerationEU).

\section{Bibliographical References}\label{sec:reference}

\bibliographystyle{lrec2026-natbib}
\bibliography{lrec2026-example}

\bibliographystylelanguageresource{lrec2026-natbib}

\appendix
\section{Appendix}
\label{appendix}

\subsection{Additional Dataset Details}
\label{appendix:additional_data_details}
In this subsection we report additional details on the datasets employed in this work. In Table~\ref{tab:fineweb_data} and Table~\ref{tab:stack_data} we present the details of the web and code datasets used for pre-training. Then, in Table~\ref{tab:ordered_language_data} we report the details for the high-quality datasets by language, while the full list of high-quality datasets with their licenses can be found in Table~\ref{tab:datasets}.
Finally, in Table~\ref{tab:messages_by_language}, the details for the pre-existing supervised fine-tuning datasets.

For our \textbf{multilingual question-answering (QA) dataset}, we sourced QA examples from a range of datasets, including grade-school science questions in English from AI2 ARC\footnote{\url{https://huggingface.co/datasets/allenai/ai2_arc}}, frequently asked questions from Swedish authorities via SweFAQ\footnote{\url{https://spraakbanken.gu.se/en/resources/swefaq}}, a Czech translation of the TruthfulQA dataset\footnote{\url{https://huggingface.co/datasets/CIIRC-NLP/truthful_qa-cs}}, and multilingual QA resources from CohereLabs Include-base-44\footnote{\url{https://huggingface.co/datasets/CohereLabs/include-base-44}} and ExamsQA\footnote{\url{https://github.com/mhardalov/exams-qa}}.

\section{Scaling Law Fitting Procedure}
\label{app:scale}

To estimate a scaling law for the loss function of our final model, we trained several smaller models with a similar protocol of the final run, and then fitted a power law to them.

The models varied in terms of parameter count (26.5M, 62.9M, and 110.6M) and effective dataset sizes ($5.8 \times 10^{10}$, $8.9 \times 10^{10}$, and $1.1 \times 10^{11}$ tokens). Each model was trained under the same WSD learning rate regime, using the same data mixtures of the final model. Notably, this includes a decay stage for the last 10\% of each training run, using the mixture with upsampled high-quality data described in the main text. 
For each configuration, we recorded the final validation loss after the decay stage, and used those values as fitting targets.

We fitted the parameters $(L_{\infty}, A_N, \alpha, A_D, \beta)$ using nonlinear least squares optimization with the \texttt{scipy.optimize.curve\_fit} function. Both model size $N$ and dataset size $D$ were provided as joint inputs, allowing a simultaneous fit of the two scaling dimensions. We constrained the parameters to physically meaningful ranges:
\begin{itemize}[noitemsep,topsep=0pt]
    \item $L_{\infty} \in [1, 10]$
    \item $A_N, A_D > 100$
    \item $\alpha, \beta \in [0.3, 0.5]$
\end{itemize}

\begin{figure}[t!]
    \centering
    \includegraphics[width=\linewidth]{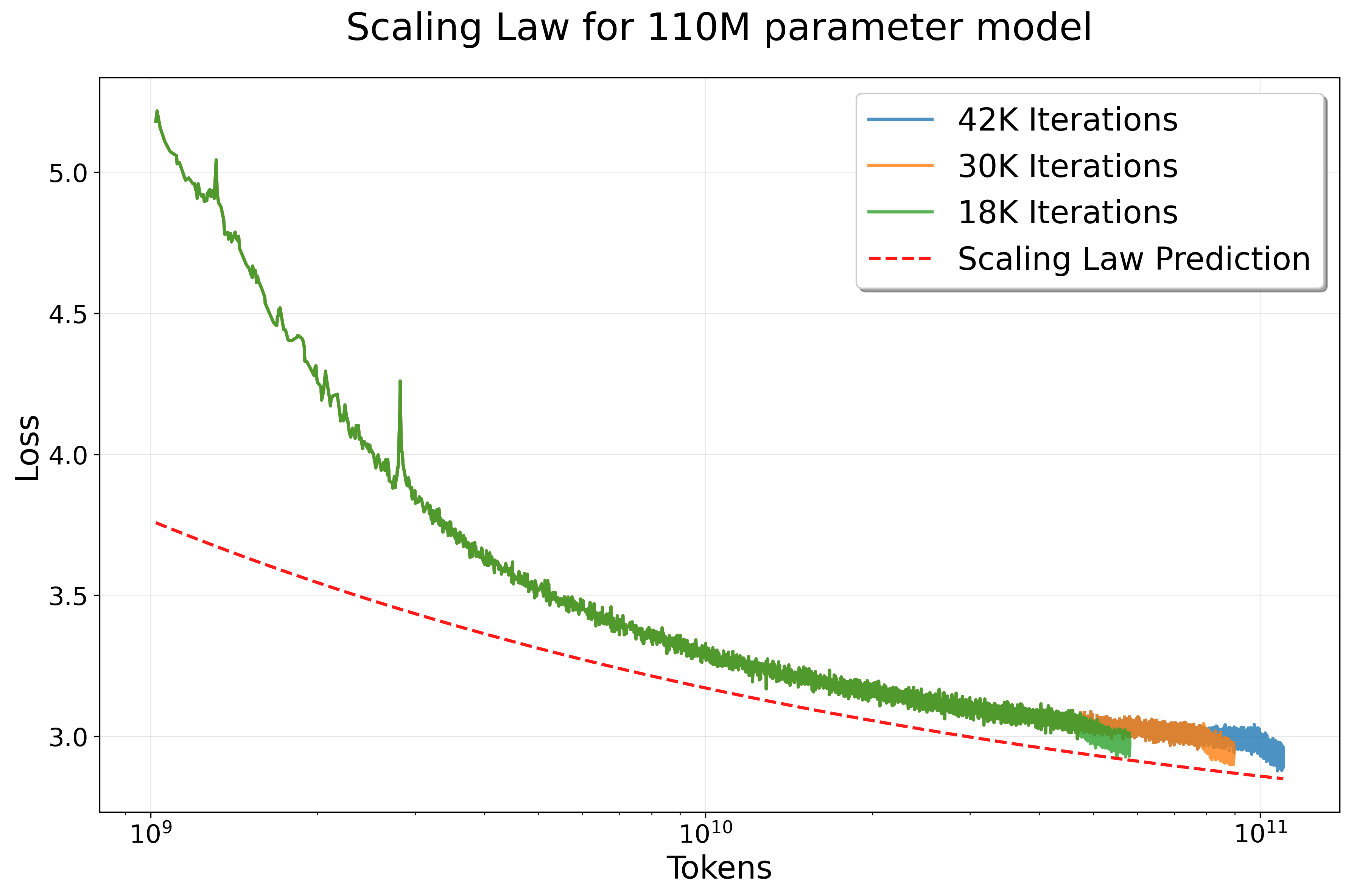}
    \caption{Example of scaling law fitting for 110M parameter model in logarithmic scale.}
    \label{fig:scale_110}
\end{figure}

The fitting was initialized with a moderate learning rate and allowed up to 30,000 function evaluations to ensure convergence. The example of the fitted scaling law for 110M model based on observations is illustrated in Figure~\ref{fig:scale_110}.

While the overall fit captured the empirical trend, the limited number of data points and narrow model size range naturally constrain precision. Increasing the diversity of training runs (especially larger models and longer training schedules) would further improve the robustness of the estimated exponents $\alpha$ and $\beta$.

\section{Decay stage data mixture}
\label{app:mix}

We experimented with varying percentage of high quality data and upsampling of low resource languages by repeating samples. As shown in Table~\ref{tab:decay_data_distribution}, the best result is obtained by increasing the high-quality data share, while not changing the language distribution. All experiments were conducted for one billion parameter model.

\begin{table}[h!]
\centering
\caption{\textbf{Data mixtures for decay stage.} Effect of data distribution strategies on validation loss and perplexity.}
\label{tab:decay_data_distribution}
\begin{tabular}{lcc}
\toprule
\textbf{Sampling strategy} & \textbf{\thead{High-quality \\ data (\%)}} & \textbf{\thead{Test \\ loss}}  \\
\midrule
Pre-train distribution       & 10 & 2.30  \\
Upsampled low-resource     & 10 & 2.28  \\
Pre-train distribution     & 30 & \textbf{2.16} \\
Upsampled low-resource     & 30 & 2.17 \\
\bottomrule
\end{tabular}
\end{table}

\clearpage
\begin{table}[t]
\centering
\caption{\textbf{Pre-training web data details.} FineWeb-2 token count and percentage share (over total web data) for each language before cleaning.}
\label{tab:fineweb_data}
\begin{tabular}{@{}lrr@{}}
\toprule
\textbf{Language} & \textbf{Token Count} & \textbf{Percentage} \\
\midrule
English    & 325B & 30.69\% \\
German     & 213B & 20.12\% \\
Spanish    & 177B& 16.73\% \\
French     & 177B & 16.75\% \\
Italian    & 78B  & 7.41\% \\
Dutch      & 62B  & 5.90\% \\
Portuguese & 107B & 10.11\% \\
Polish     & 50B  & 4.76\% \\
Swedish    & 22B & 2.16\% \\
Czech      & 30B  & 2.86\% \\
Finnish    & 20B & 1.89\% \\
Bulgarian  & 11B  & 1.08\% \\
Greek      & 12B & 1.18\% \\
\midrule
\textbf{Total} & 1{,}292B & 100\% \\
\bottomrule
\end{tabular}
\end{table}

\begin{table}[b]
\centering
\caption{\textbf{Pre-training code data details.} Token count and percentage share (over total code data) for the selected programming languages from the Stack dataset.}
\label{tab:stack_data}
\begin{tabular}{@{}lrr@{}}
\toprule
\textbf{Language} & \textbf{Token Count} & \textbf{Percentage} \\
\midrule
C         & 5B & 10.00\% \\
JSON      & 12B & 25.00\% \\
Markdown  & 7B & 15.00\% \\
Python    & 5B  & 10.00\% \\
Shell     & 1B  & 2.00\% \\
SQL       & 5B  & 10.00\% \\
TeX       & 4B & 8.00\% \\
YAML      & 5B & 10.00\% \\
XML       & 5B & 10.00\% \\
\midrule
\textbf{Total} & 50B & 100\% \\
\bottomrule
\end{tabular}
\end{table}

\begin{table}[t]
\centering
\caption{\textbf{High-quality pre-training data details.} Token counts (rounded to billions) and percentage shares (over total high-quality data) for each language.}
\begin{tabular}{lrr}
\toprule
\textbf{Language} & \textbf{Token Count} & \textbf{Percentage} \\
\midrule
English     & 71B & 34.50\% \\
German      & 51B & 24.64\% \\
Spanish     & 19B & 9.43\% \\
French      & 26B & 12.84\% \\
Italian     & 10B & 4.83\% \\
Dutch       & 3B  & 1.50\% \\
Portuguese  & 6B  & 2.88\% \\
Polish      & 5B  & 2.42\% \\
Swedish     & 3B  & 1.22\% \\
Czech       & 5B  & 2.24\% \\
Finnish     & 2B  & 0.84\% \\
Bulgarian   & 2B  & 1.14\% \\
Greek       & 3B  & 1.52\% \\
\midrule
\textbf{Total} & 206B & 100.00\% \\
\bottomrule
\end{tabular}
\label{tab:ordered_language_data}
\end{table}

\begin{table}[b]
\centering
\caption{\textbf{Supervised fine-tuning data details.} Number of instructions, relative percentage share and token count for each language.}
\label{tab:messages_by_language}
\resizebox{\linewidth}{!}{
\begin{tabular}{lrrr}
\toprule
\textbf{Language} & \textbf{Instruction} & \textbf{Percentage} & \textbf{Token} \\
 & \textbf{Count} & & \textbf{Count} \\
\midrule
Spanish     & 83,753  & 6.52\%  & 10M \\
English     & 300,113 & 23.38\% & 47M \\
German      & 98,150  & 7.65\%  & 18M \\
French      & 90,247  & 7.03\%  & 16M \\
Finnish     & 99,795  & 7.77\%  & 13M \\
Polish      & 84,938  & 6.61\%  & 11M \\
Portuguese  & 82,334  & 6.41\%  & 11M \\
Italian     & 84,684  & 6.60\%  & 11M \\
Bulgarian   & 86,279  & 6.72\%  & 13M \\
Czech       & 95,872  & 7.47\%  & 15M \\
Greek       & 67,789  & 5.28\%  & 10M \\
Dutch       & 80,714  & 6.29\%  & 11M \\
Swedish     & 78,930  & 6.14\%  & 10M \\
Code        & 50,644  & 3.94\%  & 20M \\
\midrule
\textbf{Total} & 1,3M & 100.00\% & 229M \\
\bottomrule
\end{tabular}
}
\end{table}

\begin{table*}[h]
\centering
\caption{\textbf{Model configurations.} Model configurations used for MiniLingua in the scaling law experiments and in the final model.}
\label{tab:scaling_configs}
\resizebox{\textwidth}{!}{
\begin{tabular}{lcccccccc}
\toprule
\textbf{Model} & \textbf{Layers} & \textbf{Model Dim} & \textbf{FFN Dim} & \textbf{Heads} & \textbf{Seq Len} & \textbf{KV Heads} & \textbf{Head Dim} & \textbf{Params (non-emb)} \\
\midrule
MiniLingua-30M  & 12 & 384 & 1536 & 6  & 2048 & 3 & 64 & 26.6M \\
MiniLingua-60M  & 16 & 512 & 2048 & 8  & 2048 & 4 & 64 & 62.9M \\
MiniLingua-110M & 18 & 640 & 2560 & 10 & 2048 & 5 & 64 & 110.6M \\
MiniLingua-1B   & 32 & 1536 & 6144 & 24 & 2048 & 8 & 64 & 1.11B \\
\bottomrule
\end{tabular}
}
\end{table*}


\begin{table*}[ht]
\centering
\caption{\textbf{High-quality datasets and licenses.} Dataset name and license for the 35 datasets used in our high-quality multi-lingual dataset for the last phase of pre-training.}
\label{tab:datasets}
\begin{tabular}{ll}
\toprule
\textbf{Dataset} & \textbf{License} \\
\midrule




EMEA-V3 & Attribution 4.0 International \\
Europarl & No known copyright restrictions \\
Opus Books & Copyright-free books \\
Eurovoc & Creative Commons CC0 \\
News Commentary & Apache License 2.0 \\
Eac\_tm & Creative Commons Attribution 4.0 (CC BY 4.0) \\
Academic Texts CLARIN & MIT License \\
Opus\_100 & MIT License \\
Chitanka (HF) & Apache 2.0 \\
MOSEL & CC BY 4.0 \\
Wikipedia & CC BY-SA 3.0 \\
News & CC BY-NC-ND 4.0 \\
BNC (1951--2021) & CC BY-NC-SA 3.0 \\
CS Academic Abstracts & CC BY 4.0 \\
Academic Texts & CC BY 4.0 \\
Literaturbanken & CC BY 4.0 \\
Poems & CC BY 4.0 \\
SVT News 2022 & CC BY 4.0 \\
CLEAR Drugs and Notes & CC BY 4.0 \\
Multilingual Medical Corpus & Apache 2.0 \\
OpenSubtitles & Apache 2.0 \\
Academic Texts 2 & CC BY 4.0 \\
Swedish Books & CC BY 4.0 \\
WMT17 & CC BY 4.0 \\
Yle News & CC BY \\
CurliCat Polish Corpus (ELRC-SHARE) & CC BY-SA 4.0 \\
Clarin-PL Corpus & CC BY 4.0 \\
European Language Grid Corpus & CC BY 4.0 \\
Czech SYN2015 & Academic use \\
ParlSpeechv2 & CC0 1.0 \\
Ylilauta (Kielipankki) & CC BY-NC \\
JRC-Acquis & Public domain (JRC-Acquis license) \\
Suomi24 (2018-2020) & ACA+NC \\
BG News & CC BY 4.0 \\
OpenCulture & CC0 1.0 \\
\bottomrule
\end{tabular}
\end{table*}

\end{document}